\documentclass[10pt,twocolumn,letterpaper]{article}

\usepackage[pagenumbers]{author-kit/cvpr} 
\usepackage{multirow}

\usepackage[normalem]{ulem}

\definecolor{cvprblue}{rgb}{0.21,0.49,0.74}
\usepackage[pagebackref,breaklinks,colorlinks,allcolors=cvprblue]{hyperref}

\def\paperID{******} 
\def\confName{CVPR}
\def\confYear{2026}

\title{VGGT4D: Mining Motion Cues in Visual Geometry Transformers\\ for 4D Scene Reconstruction}

\author{
Yu Hu\textsuperscript{1}\thanks{Equal contribution.} \quad
Chong Cheng\textsuperscript{1,2}\footnotemark[1] \quad
Sicheng Yu\textsuperscript{1}\footnotemark[1] \\
Xiaoyang Guo\textsuperscript{2} \quad
Hao Wang\textsuperscript{1}\thanks{Corresponding author.} \\
\textsuperscript{1}The Hong Kong University of Science and Technology (Guangzhou) \\
\textsuperscript{2}Horizon Robotics \\
}

\begin{document}
\maketitle

\begin{abstract}

    Reconstructing dynamic 4D scenes is challenging, as it requires robust disentanglement of dynamic objects from the static background. While 3D foundation models like VGGT provide accurate 3D geometry, their performance drops markedly when moving objects dominate. Existing 4D approaches often rely on external priors, heavy post-optimization, or require fine-tuning on 4D datasets. 
    In this paper, we propose VGGT4D, a training-free framework that extends the 3D foundation model VGGT for robust 4D scene reconstruction. Our approach is motivated by the key finding that VGGT's global attention layers already implicitly encode rich, layer-wise dynamic cues. To obtain masks that decouple static and dynamic elements, we mine and amplify global dynamic cues via gram similarity and aggregate them across a temporal window. To further sharpen mask boundaries, we introduce a refinement strategy driven by projection gradient. We then integrate these precise masks into VGGT's early-stage inference, effectively mitigating motion interference in both pose estimation and geometric reconstruction.
    Across six datasets, our method achieves superior performance in dynamic object segmentation, camera pose estimation, and dense reconstruction. It also supports single-pass inference on sequences longer than 500 frames.
    \\Project Page: \url{https://3dagentworld.github.io/vggt4d/}

\end{abstract}
\section{Introduction}

Reconstructing 4D scenes with dynamic objects from visual inputs has been a challenging task. This is because moving objects not only degrade pose estimation but also interfere with the background geometry modeling, and their motion is often entangled with camera motion, which leads to severe artifacts in 3D scene representations. Therefore, how to model dynamics is crucial for robust 4D reconstruction.

\begin{figure}[t]
  \begin{center}
  \includegraphics[width=1.0\linewidth]{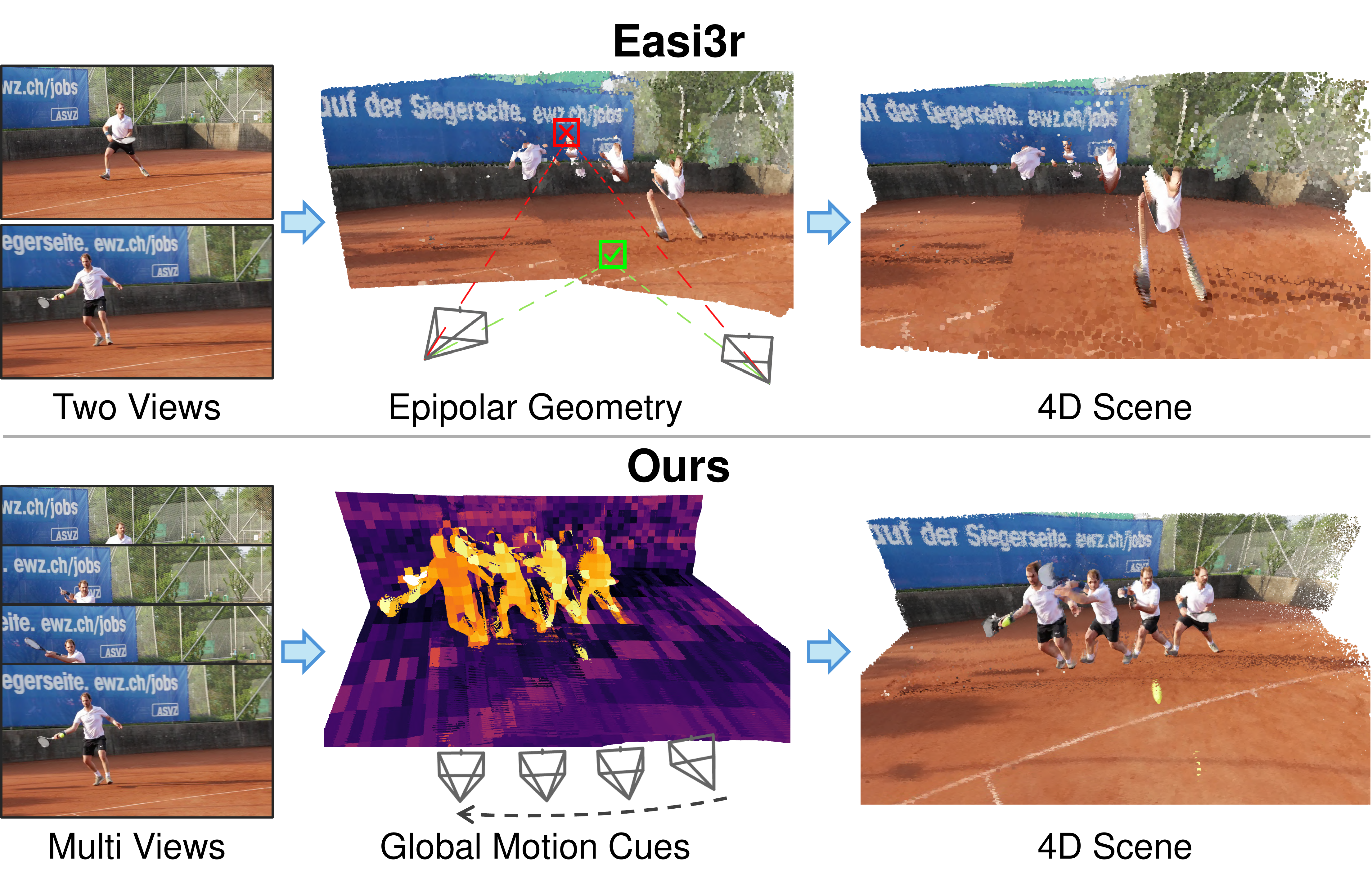}
  \end{center}
  \vspace{-15pt}
  \caption{
  Easi3R, built upon DUSt3R, is restricted to two-view inputs and derives dynamic region masks by identifying epipolar-inconsistent pixels.
  In contrast, our proposed VGGT4D reconstructs dynamic scenes from multi-view inputs by extracting global motion cues from VGGT’s attention maps. 
  }
  \vspace{-15pt}
\end{figure}

Traditional SfM~\cite{agarwal2011sfm,schonberger2016structure-sfm2} and MVS \cite{schonberger2016mvs1,furukawa2009mvs2} methods rely on multi-view rigidity and photometric constancy. Dynamic regions violate these assumptions, degrading correspondences and bundle adjustment, and often causing failure. 3D foundation models such as VGGT~\cite{wang2025vggt} deliver fast, accurate 3D geometry and camera pose estimation, yet they are largely trained and inferred under static-scene assumptions and lack an explicit mechanism to disentangle moving objects. When dynamics dominate, this coupling of dynamics and statics leads to brittle reconstruction and pose drift.

While existing methods have made progress in 4D reconstruction, they suffer from two limitations: (1) heavy iterative refinement that incurs substantial runtime and memory overhead~\cite{kopf2021robust-CVD,zhang2022Casual-SAM,yao2025uni4d}; (2) reliance on external modules (optical flow~\cite{zhang2024monst3r}, depth~\cite{li2025megasam}, semantic segmentation~\cite{goli2025romo}), which complicates integration and makes performance sensitive to module quality and domain shift. Recent efforts have explored efficient feed-forward architectures~\cite{zhang2024monst3r,xu2024das3r,cut3r,xiao2025spatialtrackerv2,zhang2025pomato,jiang2025geo4d}, but most still require large-scale training or fine-tuning on high-quality dynamic datasets, which are costly to curate and scarce at scale.

To alleviate the issues above, this paper aims to discover: 
\textit{can we endow 4D reconstruction capability to a 3D foundation model—without additional training?} 

A preliminary step toward this goal is Easi3R~\cite{chen2025easi3r}, a training-free extension of DUSt3R~\cite{dust3r_cvpr24} that segments dynamic masks by analyzing the spatial and temporal statistics of decoder attention. 
However, Easi3R is built upon a pairwise cross-attention architecture, which captures only local feature interactions. 
This design imposes a short temporal horizon and yields masks that are inconsistent across frames, with boundary errors at dynamic–static interfaces that cause depth drift and floating artifacts in the reconstructed point clouds. 
Moreover, its core assumption that tokens violating epipolar geometry receive low attention does not generalize to VGGT, whose global attention aggregates signals across multiple views.

In this paper, we propose \textbf{VGGT4D}, which extends a pretrained VGGT model to 4D scene reconstruction without further retraining. 
Our design is motivated by empirical evidence of a consistent layer-wise trend in the original VGGT: shallow transformer layers capture salient motion information, which gradually fade in deeper layers.

Unlike Easi3R’s pairwise attention statistics, our multi-frame, layer-aware attention mining generalizes to VGGT and yields globally consistent, robust dynamic masks.

Specifically, we first derive per-frame dynamic masks by aggregating gram similarity statistics across selected shallow, middle, and deep layers and a temporal window, forming a dynamic saliency signal. This signal is then refined by a projection gradient-aware strategy, which yields sharp and robust masks that can explicitly decouple dynamic and static regions. Finally, during inference, we apply these masks by suppressing dynamic image tokens only in the shallow layers, which mitigates motion interference and yields undisturbed pose estimation and 4D reconstruction.

Experiments on six dynamic datasets demonstrate superior performance in dynamic-object segmentation, camera pose estimation, and dense point-cloud reconstruction, with the ability to process sequences of more than 500 frames in a single pass.
The main contributions of this paper are summarized as follows:
\begin{enumerate}
    \item \textbf{Training-free 4D perception for VGGT.} We mine motion cues latent in VGGT’s global attention to endow a 3D foundation model with 4D perception, without additional training.
    \item \textbf{Consistent dynamic-static decoupling pipeline.} We introduce a novel method that aggregates Gram similarity statistics from VGGT's attention and sharpens the resulting saliency signals with gradient-aware refinement, yielding masks that stabilize 4D reconstruction.
    \item \textbf{Superior Performance and Generalization.} 
    Our approach outperforms existing models on six dynamic datasets across dynamic segmentation, pose estimation, and 4D reconstruction. It also successfully processes long sequences (500+ frames) in a single pass.
\end{enumerate}

\section{Related Work}

\noindent \textbf{3D Foundation Models.}
DUSt3R~\cite{dust3r_cvpr24} pioneered large-scale pretraining for feed-forward 3D foundation models by predicting dense pointmaps from image pairs. MASt3R~\cite{leroy2024mast3r} strengthened correspondence quality, and Reloc3R~\cite{dong2025reloc3r} directly regressed 6-DoF poses to sharpen camera estimation. However, pairwise inputs make inference cost grow quadratically with sequence length. To lift this constraint, multi-image variants introduce memory encoders~\cite{wang2024spann3r,cabon2025must3r} and subgraph fusion~\cite{liu2024slam3r} to aggregate context without exhaustive pairing. VGGT~\cite{wang2025vggt} and Fast3R~\cite{Yang_2025_fast3R} further employ global attention for cross-view reasoning, while MV-DUSt3R+~\cite{tang2024mv-dust3r+} and FLARE~\cite{zhang2025flare} couple such priors with 3D Gaussian Splatting~\cite{kerbl2023-3dgs} for end-to-end reconstruction from sparse views. Recent follow-ups push beyond these methods with unified dense-geometry prediction~\cite{fang2025dens3r}, causal streaming for long sequences~\cite{lan2025stream3r,zhuo2025streamVGGT}, permutation-equivariant, reference-free design~\cite{wang2025pi}, and training-free token-merging acceleration for VGGT~\cite{shen2025fastvggt}—delivering faster inference and stronger generalization.

\noindent \textbf{4D Scene Reconstruction.}
Early methods often rely on RGB-D sensors, which limit applicability in real-world scenes~\cite{newcombe2015dynamicfusion,innmann2016volumedeform}. A large body of methods jointly estimate depth, pose, and residual motion (typically with a motion mask over stationary regions), including self-supervised variants ~\cite{gordon2019depth-in-wild, mahjourian2018unsupervised, godard2019digging, yang2018every} and test-time or optimization-based schemes such as CasualSAM~\cite{zhang2022Casual-SAM} and Robust-CVD~\cite{kopf2021robust-CVD}. More recent pipelines, e.g., MegaSaM~\cite{li2025megasam}, achieve robust poses and reconstructions on dynamic videos by leveraging strong monocular depth priors, while Uni4D~\cite{yao2025uni4d} integrates multiple visual foundation models with multi-stage bundle adjustment for high-quality 4D results. These approaches are effective but hinge on strong priors and heavy test-time optimization, which limits scalability. 
Recent works reduce or replace test-time optimization using learned priors or attention adaptation: MonST3R~\cite{zhang2024monst3r} fine-tunes on dynamic data and exploits optical flow; DAS3R~\cite{xu2024das3r} augments a DPT head for feed-forward motion masks; CUT3R~\cite{cut3r} fine-tunes MASt3R on mixed static/dynamic data for fast reconstruction; Easi3R~\cite{chen2025easi3r} adapts attention during inference to perform training-free 4D reconstruction on DUSt3R.
Despite the progress, most pipelines require fine-tuning or second-stage post-processing, and training-free variants are often tightly coupled to specific backbones and can become unstable on long sequences. 

\section{Method}

\begin{figure*}[t]
  \vspace{-5pt}
  \begin{center}
  \includegraphics[width=0.9\linewidth]{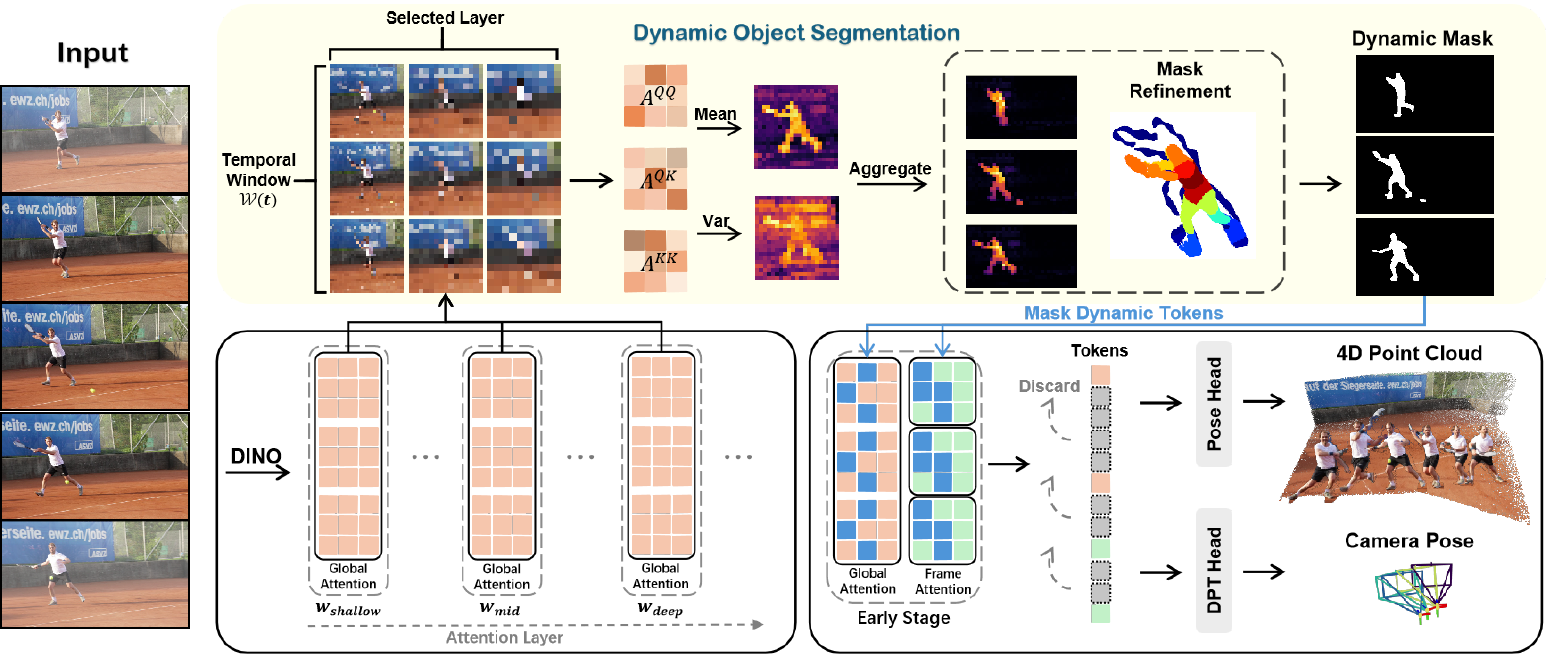}
  \end{center}
  \label{fig:framework}
  \vspace{-15pt}
  \caption{\textbf{Overview of VGGT4D.} Input image sequence is fed into VGGT. We compute and aggregate its global attention across selected layers and a temporal window to mine dynamic cues. Followed by a gradient-aware mask refinement, we get accurate dynamic masks. During inference, we apply the masks to early-stage layers and discard unused layer tokens, producing decoupled dynamic/static point clouds and camera pose estimates.}
  \vspace{-5pt}
\end{figure*}

\subsection{Preliminary}

\noindent\textbf{DUSt3R.} DUSt3R~\cite{dust3r_cvpr24} is a pretrained transformer designed for pose-free dense 3D reconstruction. It processes only two input images and performs pairwise cross-attention to compute dense pixel-level correspondences between them.

\begin{figure}
  \vspace{-10pt}
  \begin{center}
  \includegraphics[width=0.85\linewidth]{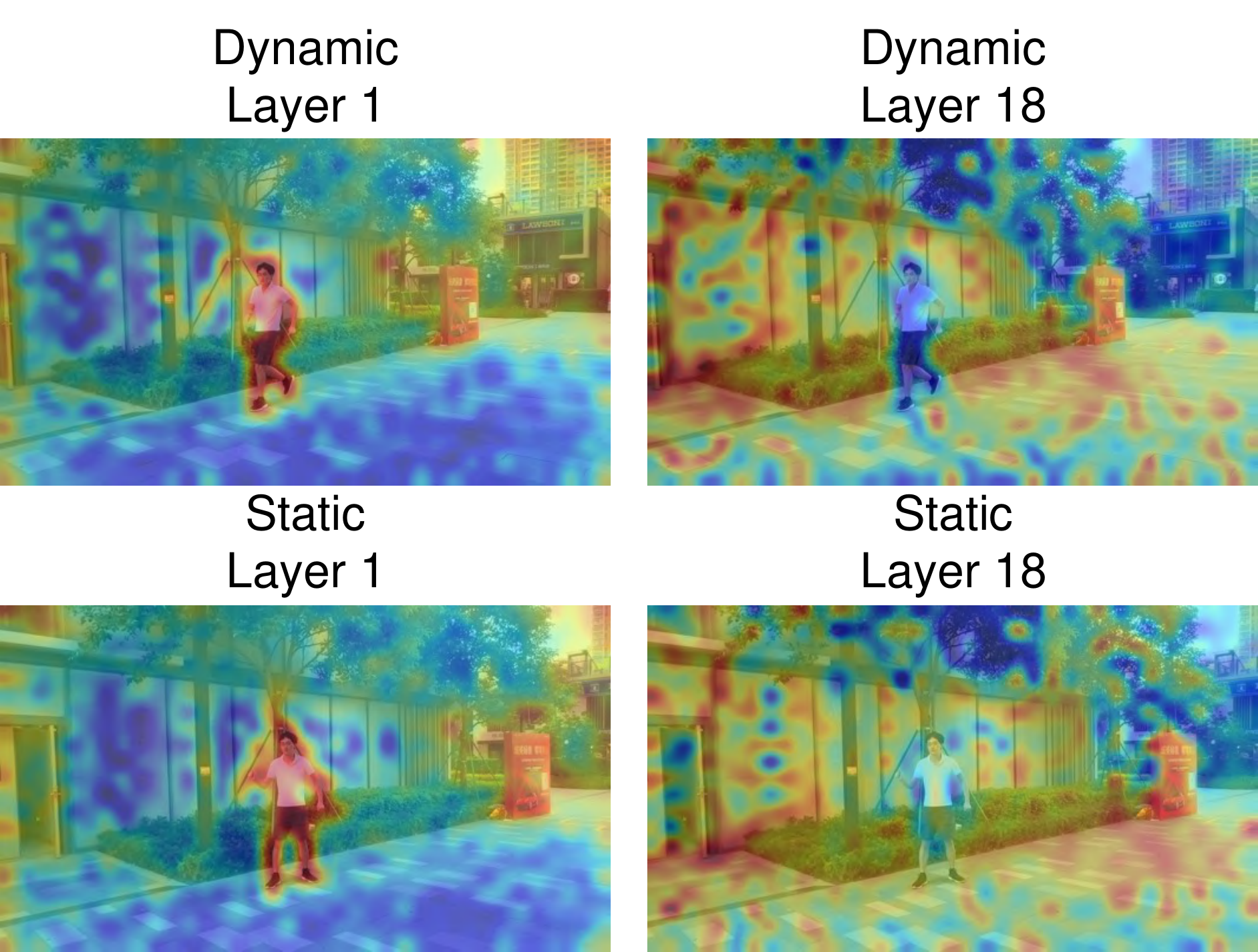}
  \end{center}
  \vspace{-5pt}
  \caption{\textbf{Visualization of VGGT's standard camera-image attention $A^{QK}$.} At layer 1, attention strongly focuses on semantic regions (e.g., people). While deeper layers can suppress physically dynamic pixels, we observe this behavior is highly scene-dependent and unreliable for robust segmentation. This limitation motivates our search for a more stable dynamic cue (\cref{sec:cue_extraction}).}
  \label{fig:vis_cam_attn}
  \vspace{-10pt}
\end{figure}

\noindent\textbf{Easi3R.} Easi3R~\cite{chen2025easi3r} extends DUSt3R to the task of 4D reconstruction. It leverages the empirical observation that DUSt3R assigns low attention to epipolar-inconsistent pixels, enabling Easi3R to derive dynamic region masks in an unsupervised manner from attention maps.

\noindent\textbf{VGGT.} VGGT~\cite{wang2025vggt} is an enhanced version of DUSt3R that processes multiple input images and jointly estimates multi-view poses and dense geometry. It employs both inter-frame and intra-frame attention mechanisms, enabling tokens to aggregate semantically coherent evidence across frames for more consistent 3D reconstruction.

\noindent\textbf{Motivation of our proposed VGGT4D.} However, since Easi3R is built upon the DUSt3R framework, it is limited to pairwise image inputs and struggles to handle multi-view inputs effectively. To overcome this limitation, we propose VGGT4D, which extends VGGT from static multi-view reconstruction to 4D dynamic scene reconstruction.

\subsection{Empirical Observations in VGGT}

We first examine whether VGGT can perceive dynamic pixels. As illustrated in ~\cref{fig:vis_cam_attn}, we visualize the attention maps between the camera token and image tokens. The results reveal a clear separation between static and dynamic regions: shallow layers respond strongly to semantically dynamic objects (e.g., people), while deeper layers gradually suppress pixels with inconsistent multi-view geometry. These results suggest that VGGT is sensitive to physical motion and implicitly encodes dynamic cues.

However, this phenomenon is highly scene-dependent and occurs only at certain tokens and layers. Meanwhile, the attention maps among image tokens contain substantial high-level semantic noise, making direct extraction unreliable. To consistently and robustly identify dynamic object masks, we propose the following pipeline to achieve 4D reconstruction in a training-free manner.

\begin{figure*}
  \vspace{-10pt}
  \begin{center}
  \includegraphics[width=0.9\linewidth]{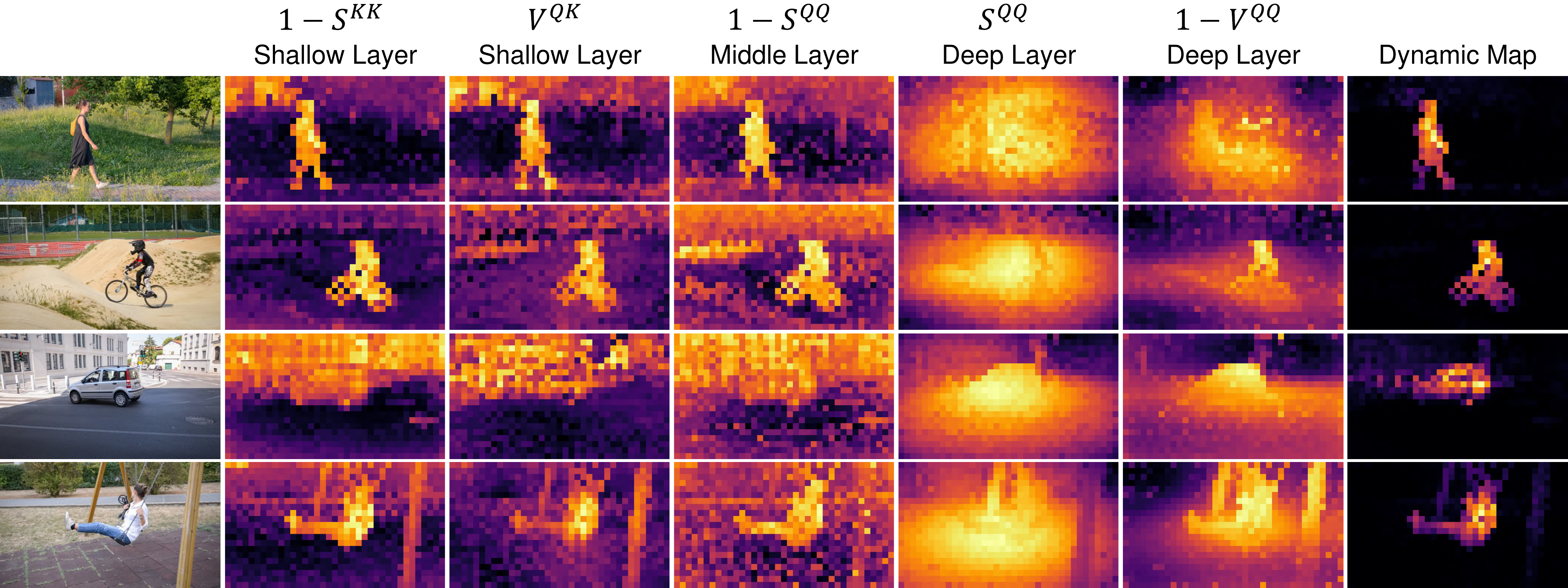}
  \end{center}
  \vspace{-5pt}
  \caption{\textbf{Visualization of gram similarity.} We visualize each component of $w_\text{shallow}$, $w_\text{middle}$ and $w_\text{deep}$ across different layers, demonstrate their complementary roles in extracting dynamic cues.}
  \label{fig:vis_attn}
  \vspace{-10pt}
\end{figure*}


\subsection{Dynamic Cue Extraction via Gram Similarity}
\label{sec:cue_extraction}
Easi3R~\cite{chen2025easi3r} extracts dynamic masks directly from the standard attention map, given by
\begin{equation}
A^{QK}_{l,t,s} = \frac{Q_{l,t} K_{l,s}^\top}{\sqrt{c}} \in \mathbb{R}^{N_p \times N_p},
\end{equation}
where $c$ is the feature dimension, $l$ indexes the layer, $t$ and $s$ denote the frame indices, and $N_p$ is the token count.

However, we find that $A^{QK}$ inherently mixes dynamic motion with texture and semantic responses, thereby reducing the clarity of motion cues. To amplify the distribution discrepancies caused by motion, we propose to compute the Gram similarities $QQ^\top$ and $KK^\top$:
\begin{equation}
\begin{aligned}
A^{QQ}_{l,t,s} = \frac{Q_{l,t} Q_{l,s}^\top}{\sqrt{c}}, \quad A^{KK}_{l,t,s} = \frac{K_{l,t} K_{l,s}^\top}{\sqrt{c}}.
\end{aligned}
\label{eq:gram_sim}
\end{equation}
These self-similarity matrices effectively enhance dynamic cues otherwise encoded implicitly within $Q$ and $K$.

To aggregate temporal information, we employ an inter-frame sliding-window strategy across frames, defined as $\mathcal{W}(t) = \{t-n, \dots, t-1, t+1, \dots, t+n\}$. Within this window and across three set of layers $L$, including shallow, middle, and deep layers, we compute the mean ($S$) and variance ($V$) of the Gram similarities:
\begin{align}
  S^{\mathrm{X}}_{i\text{--}j} \;
  &= \; \operatorname{Mean}_{s}\bigg(\;\;
  \frac{1}{|\mathcal{W}(t)|} \sum_{s\in\mathcal{W}(t)} \;
      \frac{1}{L} \sum_{l=i}^{j} \;
      A^{\mathrm{X}}_{l,t,s} \; \;
  \bigg) \, , \;
  \label{eq:gram_mean}\\
  V^{\mathrm{X}}_{i\text{--}j} \;
  &= \; \operatorname{Var}_{s}\bigg(\quad\;
  \frac{1}{|\mathcal{W}(t)|} \sum_{s\in\mathcal{W}(t)} \;
      \frac{1}{L} \sum_{l=i}^{j} \;
      A^{\mathrm{X}}_{l,t,s} \; \;
  \bigg) \, , \;
  \label{eq:gram_std}
\end{align}
where $\mathrm{X} \in \{QQ, QK, KK\}$, $i$ and $j$ denote the start and end layer indices, respectively.

Finally, we construct a dynamic saliency map $Dyn$ by mining these statistics:
\begin{equation}
  \mathrm{Dyn} = w_{\text{shallow}} \odot w_{\text{middle}} \odot w_{\text{deep}},
  \label{eq:dyn}
\end{equation}
where $\odot$ denotes element-wise multiplication. The three factors capture complementary cues:
\begin{align}
    w_{\text{shallow}} &= (1 - S^{KK}_{\text{shallow}}) \odot V^{QK}_{\text{shallow}},
    \label{eq:w_sem}\\
    w_{\text{middle}} &= 1 - S^{QQ}_{\text{middle}},
    \label{eq:w_mid}\\
    w_{\text{deep}} &= (1 - V^{QQ}_{\text{deep}}) \odot S^{QQ}_{\text{deep}}.
    \label{eq:w_deep}
\end{align}
As illustrated in \cref{fig:vis_attn}, $w_{\text{shallow}}$ (shallow layers) captures semantic saliency, $w_{\text{middle}}$ (middle layers) identifies motion instability, and $w_{\text{deep}}$ (deep layers) acts as a spatial prior to suppress outliers. The per-frame dynamic mask is then obtained by thresholding, $M_t = [\mathrm{Dyn} > \alpha]$, and refined with feature clustering (see supplementary).

\begin{table*}[h]
\vspace{-5pt}
\centering
{
\small
\renewcommand{\arraystretch}{0.9}
\begin{tabular}{l|cccc|cccc|cccc}
\toprule
\multirow{2}{*}{\textbf{Method}} & \multicolumn{4}{c|}{DAVIS-2016} & \multicolumn{4}{c|}{DAVIS-2017} & \multicolumn{4}{c}{DAVIS-all} \\ \cmidrule(lr){2-5} \cmidrule(lr){6-9} \cmidrule(lr){10-13}
                               & JM↑ & JR↑ & FM↑ & \multicolumn{1}{c|}{FR↑} & JM↑ & JR↑ & FM↑ & \multicolumn{1}{c|}{FR↑} & JM↑ & JR↑ & FM↑ & FR↑ \\ \midrule
$\text{Easi3R}_\text{dust3r}$  & 50.10 & 55.77 & 43.40 & 37.25 & 46.86 & 50.54 & 39.06 & 30.05 & 44.10 & 50.85 & 35.16 & 27.24 \\
$\text{Easi3R}_\text{monst3r}$ & \underline{54.93} & \underline{68.00} & 45.29 & 47.30 & \underline{54.75} & \textbf{66.16} & 44.09 & 42.36 & \textbf{51.64} & \textbf{63.06} & 40.98 & 38.49 \\
MonST3R                        & 40.42 & 40.39 & \underline{49.54} & \underline{52.12} & 38.07 & 36.05 & \underline{48.24} & \underline{49.01} & 36.98 & 34.52 & \underline{47.03} & \textbf{46.72} \\
DAS3R                          & 41.13 & 38.67 & 44.50 & 36.94 & 44.51 & 43.95 & 46.71 & 44.96 & 43.33 & 38.93 & 45.24 & 38.78 \\ \midrule
Ours                           & \textbf{62.12} & \textbf{76.80} & \textbf{56.04} & \textbf{67.49} & \textbf{56.45} & \underline{65.62} & \textbf{51.09} & \textbf{56.85} & \underline{50.75} & \underline{55.59} & \textbf{47.04} & \underline{46.43} \\ \bottomrule
\end{tabular}
}
\vspace{-5pt}
\caption{\textbf{Dynamic object segmentation result on DAVIS dataset.} The best and second best results are \textbf{bold} and \underline{underlined}, respectively. The subscript \textit{DUSt3R}/\textit{MonST3R} in Easi3R denotes the variant using the DUSt3R and MonST3R backbone.}
\label{tab:mask_est}
\vspace{-5pt}
\end{table*}

\begin{figure*}
\vspace{-5pt}
\begin{center}
\includegraphics[width=0.85\textwidth]{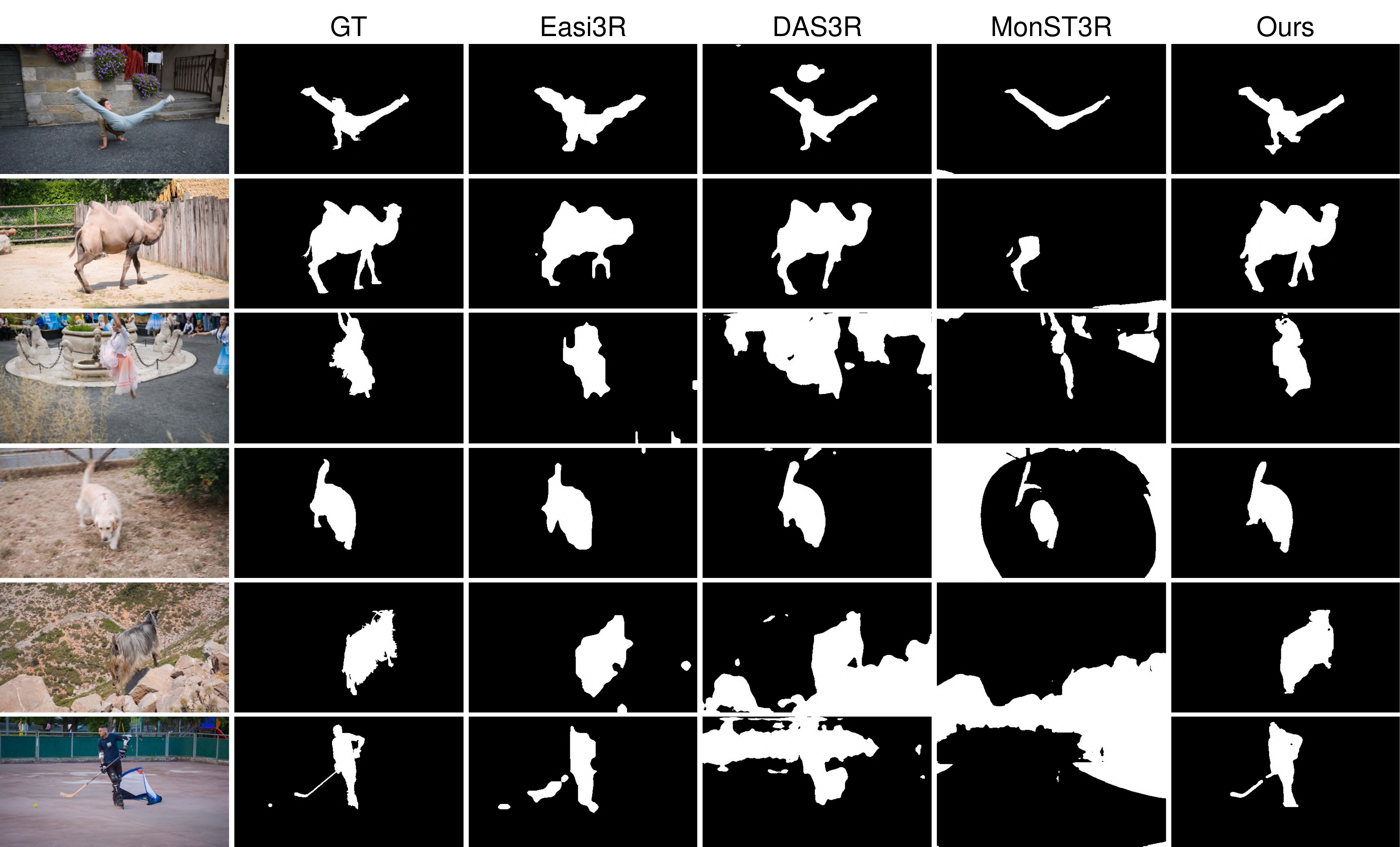}
\end{center}
\vspace{-10pt}
\caption{\textbf{Qualitative results of dynamic object segmentation.} Our method extracts sharp and accurate masks. In contrast, baseline methods suffer from coarse boundaries, missed details, and significant over-segmentation.}
\label{fig:davis_mask}
\vspace{-10pt}
\end{figure*}

\subsection{Mask Refinement via Projection Gradients}
\label{sec:mask_refinement}

The masks extracted from ViT layers are coarse and lead to ``floaters" in 4D reconstruction. Therefore, we propose a refinement method driven by point-cloud projection gradients to improve boundary accuracy.

Our core idea is that 3D points from dynamic objects will have large geometric and photometric errors when projected onto the \textit{static} regions of other views. We first define a geometric loss for a 3D point projected onto view $i$:
\begin{equation}
\begin{aligned}
    \mathcal{L}_{proj} &= \frac{1}{2} \, \mathbb{I}_i \, \bigl(1-M_i\bigr) \, \bigl\| r_{d,i} \bigr\|^2_2 \;,
\end{aligned}
\end{equation}
where $r_{d,i} = d_i - D_i(u_i,v_i)$, $M_i$ is the initial dynamic mask, $\mathbb{I}_i$ is the visibility mask, and $r_{d,i}$ is the depth residual between the projected depth $d_i$ and the depth map $D_i$.

Points with large geometric gradients are likely dynamic. We compute the gradient of this residual $r_{d,i}$ with respect to the 3D point's coordinates. It is notable this gradient $\nabla r_{d,i}$ depends on the projection Jacobians and the spatial gradient of the target depth map.

To get a robust score, we aggregate these projection gradients across all $N$ views:
\begin{equation}
\begin{aligned}
    \mathbf{agg}^{\text{proj}} &= \frac{1}{N} \sum^{N}_{i} \bigl\| w_i \, r_{d,i} \, \nabla r_{d,i} \bigl\| \; ,\\
    \text{where}& \quad w_i = \mathbb{I}_i \, \bigl(1-M_i\bigr). 
\label{eq:agg_proj}
\end{aligned}
\end{equation}

While the geometric gradient is effective, it can be unreliable in textureless regions (\eg, flat walls or floors) where depth gradients are uninformative. We therefore complement it with an aggregated photometric residual:
\begin{align}
\begin{split}
    \mathbf{agg}^{\text{photo}} = \frac{1}{N} \sum^{N}_{i} \bigl\| w_i \, \bigl(\,c - C_i(u_i,v_i)\,\bigr) \bigl\| \; ,
\end{split}
\label{eq:agg_photo}
\end{align}
where $c$ is the point's color and $C_i$ is the sampled color in view $i$.

Finally, we combine these scores to re-classify the point:
\begin{equation}
    \mathbf{agg}^{\mathrm{total}}=\mathbf{agg}^{\mathrm{proj}}+\lambda\mathbf{agg}^{\mathrm{photo}}.
\end{equation}
A point is labeled dynamic if $\mathbf{agg}^{\mathrm{total}} > \tau$. This gradient-aware refinement effectively sharpens mask boundaries. We also apply point cloud filtering and spatial clustering for robustness (see supplementary).

\begin{table*}[!ht]
\vspace{-5pt}
\centering
{
\small
\renewcommand{\arraystretch}{0.9}
\begin{tabular}{l|ccc|ccc|ccc}
\toprule
\multirow{2}{*}{\textbf{Method}} & \multicolumn{3}{c|}{Sintel} & \multicolumn{3}{c|}{TUM-Dynamics} & \multicolumn{3}{c}{VKITTI} \\ \cmidrule(lr){2-10}
\multicolumn{1}{c|}{} & ATE↓ & RTE↓ & \multicolumn{1}{c|}{RRE↓} & ATE↓ & RTE↓ & \multicolumn{1}{c|}{RRE↓} & ATE↓ & RTE↓ & RRE↓ \\ \midrule
$\text{Easi3R}_\text{dust3r}$  & 0.372             & 0.227             & 10.356            & 0.063             & 0.046             & 2.523             & 2.789             & 0.506             & 0.108             \\
$\text{Easi3R}_\text{monst3r}$ & 0.109             & 0.051             & 0.277             & 0.133             & 0.120             & 4.366             & 2.036             & 0.173             & 0.124             \\
MonST3R                        & 0.151             & \underline{0.034} & \underline{0.258} & 0.156             & 0.103             & 12.041            & 2.272             & 0.180             & 0.091             \\
POMATO                         & 0.557             & 0.158             & 0.878             & 0.153             & 0.097             & 11.102            & 1.377             & 0.232             & 0.119             \\
SpatialTrackerV2               & \textbf{0.073}    & 0.035             & 0.340             & 0.054             & 0.041             & 2.530             & 0.720             & 0.160             & 0.127             \\
DAS3R                          & 0.125             & \textbf{0.030}    & \textbf{0.185}    & 0.155             & 0.102             & 12.038            & 2.043             & 0.169             & 0.114             \\
CUT3R                          & 0.152             & 0.077             & 0.454             & 0.054             & 0.041             & 5.346             & 5.583             & 0.381             & 0.174             \\
VGGT                           & 0.081             & 0.045             & 0.287             & \underline{0.017} & \textbf{0.020} & \underline{0.617} & \underline{0.170} & \underline{0.065} & \textbf{0.062}    \\ \midrule
Ours                           & \underline{0.076} & 0.043             & 0.273             & \textbf{0.016}    & \textbf{0.020}    & \textbf{0.612}    & \textbf{0.164}    & \textbf{0.064}    & \textbf{0.062} \\ \bottomrule
\end{tabular}
}
\vspace{-5pt}
\caption{\textbf{Quantitative comparisons of camera pose estimation.} Our method consistently improves upon the strong VGGT baseline and achieves SOTA results on the TUM-Dynamics and VKITTI datasets.}
\label{tab:pose_est}
\vspace{-5pt}
\end{table*}

\subsection{4D Reconstruction via Early-Stage Masking}

With precise dynamic masks, we perform inference by integrating them into VGGT. This explicitly disentangles dynamic interference, enabling robust camera pose estimation and 4D point cloud reconstruction.

Dynamic pixels introduce geometric inconsistencies. However, naively masking all dynamic tokens in all layers is detrimental. VGGT already learns to partially attenuate dynamic signals; full-scale masking pushes the model into an out-of-distribution state, amplifying errors and removing valid static regions.

We propose a early-stage masking strategy. We mask dynamic image tokens \textit{only} in the shallow semantic and mid-level layers (specifically, layers 1-5, see supplementary). We achieve this by suppressing the Key ($K$) vectors of dynamic tokens in these layers. This approach prevents dynamic information from contaminating the deeper geometric inference stages, while still allowing the deeper layers to operate within their trained distribution. This controlled intervention yields stable poses and clean, disentangled dynamic and static point clouds.

\section{Experiments}

\begin{table}[]
\vspace{-5pt}
\centering
{
\small
\renewcommand{\arraystretch}{0.9}
\begin{tabular}{l|ccc}
\toprule         
\textbf{Method}             & ATE↓      & RTE↓      & RRE↓      \\ \midrule
$\text{Easi3R}_\text{dust3r}$               & -       & -       & -       \\
$\text{Easi3R}_\text{monst3r}$              & -       & -       & -       \\
MonST3R                                     & -       & -       & -       \\
POMATO                                      & -       & -       & -       \\
DAS3R                                       & -       & -       & -       \\
SpatialTrackerV2                            & -       & -       & -       \\
CUT3R                                       & 0.417             & 0.028             & 0.605   \\
FastVGGT                                    & 0.026             & 0.017             & 0.380   \\
VGGT                                        & \underline{0.022} & \underline{0.015} & \underline{0.344} \\ \midrule
Ours                                        & \textbf{0.019}    & \textbf{0.009}    & \textbf{0.290}   \\ \bottomrule
\end{tabular}
}
\vspace{-5pt}
\caption{\textbf{Quantitative comparisons of camera pose estimation on Point Odyssey dataset.} Our method achieves the best accuracy while remaining highly efficient. Many specialized 4D methods fail with an out-of-memory (OOM) error (`-').}
\label{tab:pose_est3}
\vspace{-5pt}
\end{table}

\noindent \textbf{Datasets.} We assess dynamic mask estimation on the DAVIS~\cite{Perazzi2016} dataset. Camera pose estimation is evaluated on DyCheck~\cite{gao2022dynamic}, TUM-Dynamics~\cite{sturm12iros}, Sintel~\cite{Butler:ECCV:2012}, VKITTI~\cite{cabon2020vkitti2} and Point Odyssey~\cite{zheng2023point}, where we select several highly dynamic video sequences. For DyCheck, frames are sampled every 4 steps; for TUM-Dynamics, every 30 steps. Point cloud reconstruction is further evaluated on DyCheck. To measure long-sequence reconstruction, we use Point Odyssey~\cite{zheng2023point} dataset with a sequence length of 500.

\noindent \textbf{Baselines.} We compare state-of-the-art pose-free 4D reconstruction methods, including MonST3R~\cite{zhang2024monst3r}, DAS3R~\cite{xu2024das3r}, CUT3R~\cite{cut3r}, Easi3R~\cite{chen2025easi3r}, SpatialTrackerV2~\cite{xiao2025spatialtrackerv2}, and POMATO~\cite{zhang2025pomato}. For long-sequence reconstruction, we also compare against FastVGGT~\cite{shen2025fastvggt}.

\noindent \textbf{Evaluation Metrics.} For camera pose estimation, we use ATE, RTE, RRE. For point cloud reconstruction, we follow prior work~\cite{aanaes2016large,chen2024feat2gsprobingvisualfoundation,wang2024spann3r} and measure Accuracy, Completeness, and Distance, which are all distance-based metrics. For dynamic mask estimation, we use IoU and Boundary F-measure, first averaging within each sequence and then across sequences. These metrics are denoted as JM and FM, respectively. To assess recall, we define a successful retrieval when IoU or Boundary F-measure exceeds 0.5, and report the mean recalls JR and FR.

\noindent \textbf{Implementation Details.} All experiments are conducted on a single NVIDIA A6000 GPU. We also modify VGGT’s attention operator. 
For more implementation details, please refer to the supplementary material.

\noindent \textbf{Long-sequence Inference Implementation.} To enable long-sequence inference, we build upon FastVGGT~\cite{shen2025fastvggt}. We leverage the finding that VGGT's prediction heads only consume tokens from specific layers, i.e., $(5, 12, 18, 24)$. We therefore discard all intermediate tokens from other layers during inference. This significantly reduces the memory footprint, allowing our method to process sequences of over 500 frames on Point Odyssey~\cite{zheng2023point}. Our dynamic token masking is applied on top of this efficient long-sequence backbone, bringing additional accuracy gains.

\begin{table*}[t]
\vspace{-5pt}
\centering
{
\small
\renewcommand{\arraystretch}{0.9}
\begin{tabular}{l|ccc|cc|cc|cc}
\toprule
\multirow{2}{*}{\textbf{Method}} & \multicolumn{3}{c|}{Pose Estimation} & \multicolumn{2}{c|}{Accuracy}        & \multicolumn{2}{c|}{Completeness}    & \multicolumn{2}{c}{Distance}        \\
            \cmidrule(lr){2-10}
                        & ATE↓             & RTE↓             & RRE↓                               & Mean↓ & Median↓ & Mean↓ & Median↓ & Mean↓ & Median↓ \\ \midrule
$\text{Easi3R}_\text{dust3r}$          & 0.022             & 0.009             & 0.806             & 0.070             & 0.044             & \underline{0.060} & 0.033             & 0.194             & 0.132             \\
$\text{Easi3R}_\text{monst3r}$         & 0.032             & 0.008             & 1.075             & 0.100             & 0.050             & 0.121             & 0.082             & 0.289             & 0.270             \\
MonST3R                                & 0.038             & 0.010             & 1.172             & 0.090             & 0.033             & 0.113             & 0.064             & 0.279             & 0.234             \\
POMATO                                 & 0.128             & 0.027             & 3.648             & 0.960             & 0.950             & 0.814             & 0.776             & 1.484             & 1.434             \\
SpatialTrackerV2                       & \underline{0.011} & \textbf{0.006}    & \textbf{0.347}    & 0.115             & 0.064             & 0.052             & 0.026             & 0.421             & 0.304             \\
DAS3R                                  & 0.052             & 0.012             & 1.560             & 0.192             & 0.142             & 0.250             & 0.108             & 0.428             & 0.336             \\
CUT3R                                  & 0.036             & 0.013             & 0.860             & 0.073             & 0.054             & 0.133             & 0.049             & 0.328             & 0.224             \\
VGGT                                   & 0.013             & 0.008             & 0.418             & \underline{0.028} & \underline{0.009} & 0.063             & \underline{0.019} & \underline{0.150} & \underline{0.055} \\ \midrule
Ours                                   & \textbf{0.010}    & \underline{0.007} & \underline{0.374} & \textbf{0.022}    & \textbf{0.004}    & \textbf{0.051}    & \textbf{0.012}    & \textbf{0.123}    & \textbf{0.050}    \\ \bottomrule
\end{tabular}
}
\vspace{-5pt}
\caption{\textbf{Quantitative comparisons on DyCheck dataset.} Our method achieves the best performance across all 4D reconstruction metrics. While SpatialTrackerV2 also shows strong pose results, its reconstruction quality is poor; our method excels in both tasks simultaneously.}
\label{tab:dycheck_comp}
\vspace{-5pt}
\end{table*}

\subsection{Dynamic Object Segmentation}
We first evaluate the core component of our method: dynamic object segmentation. Results are presented in ~\cref{tab:mask_est} and qualitatively in ~\cref{fig:davis_mask}.

As shown in ~\cref{tab:mask_est}, our full method significantly outperforms all other variants, achieving SOTA performance on DAVIS-2016 and DAVIS-2017 datasets. The qualitative results in ~\cref{fig:davis_mask} provide a clear picture: Easi3R's masks are coarse and miss details. DAS3R~\cite{xu2024das3r} tends to over-segment, bleeding into the static background, while MonST3R~\cite{zhang2024monst3r} often under-segments, failing to capture entire moving parts. In contrast, our method produces masks that are both more accurate and have sharper boundaries.

These results demonstrate that \textbf{our dynamic-mining is effective}. The strong performance validates our hypothesis that rich, extractable motion cues are embedded within VGGT's Gram similarity statistics, even without any 4D-specific training.

While $\text{Easi3R}_\text{monst3r}$ shows competitive recall on DAVIS-all, this stems from MonST3R's post-training on optical flow. Our method achieves its results in a training-free manner, operating solely on the pretrained VGGT model.

\begin{figure*}
  \vspace{-5pt}
  \begin{center}
  \includegraphics[width=0.85\textwidth]{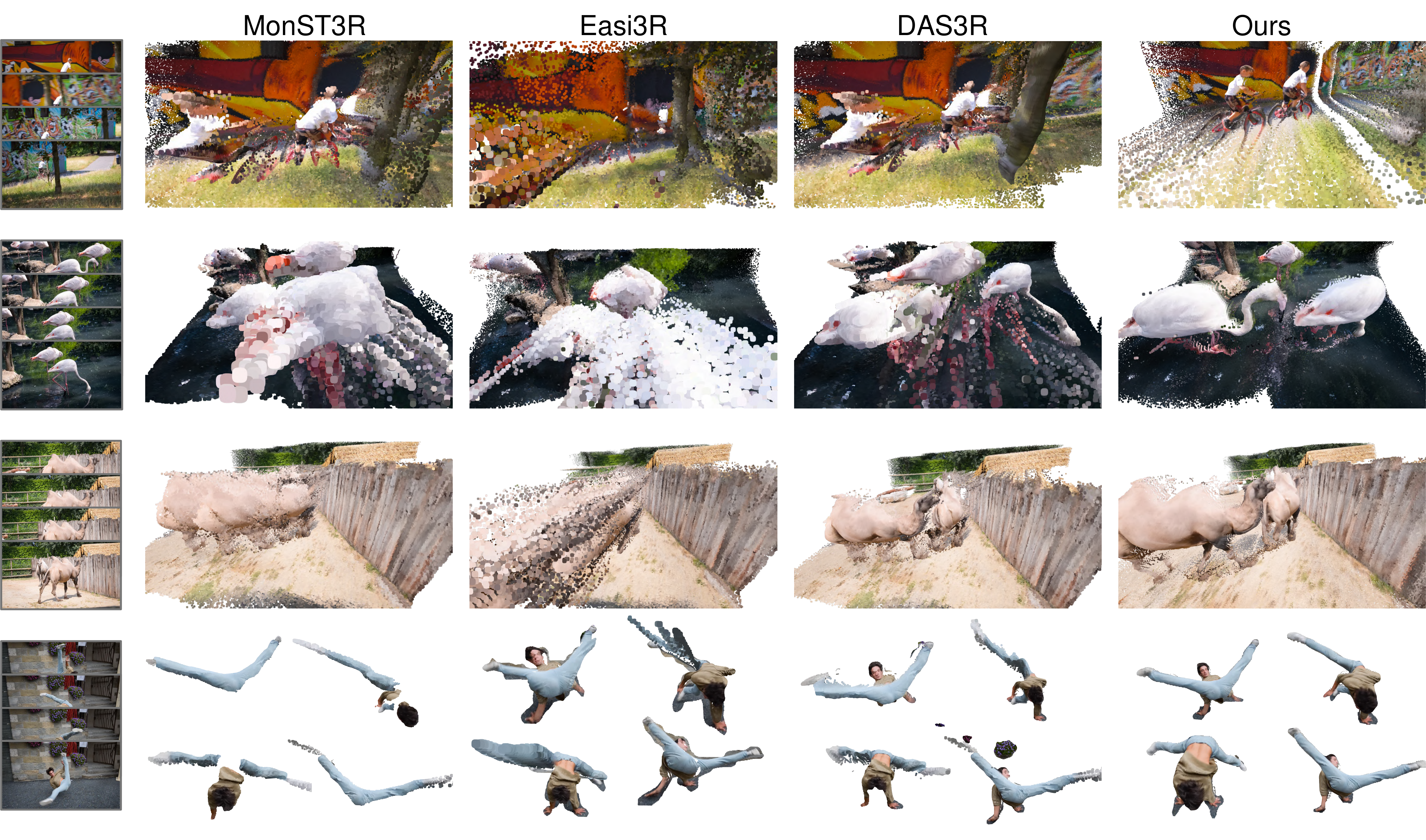}
  \end{center}
  \vspace{-10pt}
  \caption{\textbf{Qualitative comparisons of 4D reconstruction.} Baselines exhibit significant artifacts and noisy static backgrounds due to poor dynamic-static separation. In contrast, our method produces cleaner static scenes and more coherent, well-separated dynamic objects.}
  \label{fig:vis_pts}
  \vspace{-5pt}
\end{figure*}

\subsection{Camera Pose Estimation}
We evaluate camera pose estimation on several standard and challenging dynamic datasets. The results are shown in ~\cref{tab:pose_est,tab:dycheck_comp,tab:pose_est3}.

A key observation from the tables is that the original VGGT is already an exceptionally strong baseline, outperforming many specialized 4D reconstruction methods (e.g., MonST3R, DAS3R, CUT3R) on its own. This suggests its pretraining on diverse data implicitly taught it to be partially robust to dynamic objects.

However, this robustness is not perfect. Our method, VGGT4D consistently improves upon this strong VGGT baseline across all datasets. 
On the long-sequence Point Odyssey benchmark (\cref{tab:pose_est3}), we achieve the best results in all metrics while remaining highly efficient. Many other 4D methods fail to even run on this 500-frame sequence due to OOM errors.

This demonstrates that VGGT's implicit compensation is incomplete. Our explicit, training-free dynamic-static decoupling successfully identifies and removes the residual pose inconsistencies caused by motion, leading to a more stable and accurate camera trajectory, especially on long and complex sequences.

\begin{table*}[t]
\centering
{
\small
\renewcommand{\arraystretch}{0.9}
\begin{tabular}{l|cccc|cccc|cccc}
\toprule
\multirow{2}{*}{\textbf{Method}} & \multicolumn{4}{c}{DAVIS-2016} & \multicolumn{4}{|c|}{DAVIS-2017} & \multicolumn{4}{c}{DAVIS-all}     \\ \cmidrule(lr){2-13}
\multicolumn{1}{c|}{} & JM↑ & JR↑ & FM↑ & FR↑ & JM↑ & JR↑ & FM↑ & FR↑ & JM↑ & JR↑ & FM↑ & FR↑    \\ \midrule
$\text{Easi3R}_\text{vggt}$ & 7.51 & 0.12 & 12.78 & 0.00 & 10.61 & 0.08 & 16.73 & 0.08 & 10.58 & 0.12 & 17.36 & 0.24 \\
w/o refine & 59.74 & 73.10 & 50.64 & 58.30 & 54.26 & 62.72 & 46.37 & 47.32 & 47.71 & 50.39 & 40.00 & 32.40 \\ \midrule
Ours & \textbf{62.12} & \textbf{76.80} & \textbf{56.04} & \textbf{67.49} & \textbf{56.45} & \textbf{65.62} & \textbf{51.09} & \textbf{56.85} & \textbf{50.75} & \textbf{55.59} & \textbf{47.04} & \textbf{46.43} \\ \bottomrule
\end{tabular}
}
\vspace{-5pt}
\caption{\textbf{Ablation on dynamic mask estimation.} We validate our design choices: (1) Easi3R's epipolar logic fails on VGGT (\textit{$\text{Easi3R}_\text{vggt}$}), confirming its incompatibility. (2) Our core Gram similarity mining (\textit{w/o refine}) is already highly effective. (3) The gradient-based refinement (\textit{Ours}) provides a significant final boost by sharpening boundaries.}
\label{tab:ablation_mask}
\vspace{-5pt}
\end{table*}

\begin{table}[]
\vspace{-5pt}
\centering
{
\small
\renewcommand{\arraystretch}{0.9}
\begin{tabular}{l|ccc}
\toprule                     
\textbf{Method} & ATE↓    & RTE↓    & RRE↓    \\ \midrule
Full Mask                                   & 0.0302 & 0.0213 & 1.0660 \\
VGGT                                     & 0.0131 & 0.0082 & 0.4185 \\ \midrule
Ours                                        & \textbf{0.0106} & \textbf{0.0072} & \textbf{0.3746} \\ \bottomrule
\end{tabular}
}
\vspace{-5pt}
\caption{\textbf{Ablation on attention mask on DyCheck dataset.} \textit{Full Mask} denotes applying dynamic masks to all attention layers of VGGT and \textit{Ours} refers to our proposed method.}
\label{tab:ablation_decoder}
\vspace{-10pt}
\end{table}

\subsection{4D Reconstruction}
We evaluate the final 4D point cloud reconstruction quality on the DyCheck dataset (\cref{tab:dycheck_comp}) and present qualitative results in \cref{fig:vis_pts}.

Quantitatively, our method achieves the best performance on all reconstruction metrics (Accuracy, Completeness, and Distance). The improvement over the strong VGGT baseline is significant, for example, reducing the median Accuracy error from 0.009 to 0.004 and the mean Distance from 0.150 to 0.123.

These numbers are confirmed by the visual evidence in \cref{fig:vis_pts}. Baselines like MonST3R and Easi3R produce noisy and artifact-ridden reconstructions. Dynamic objects are often "ghosted" across the static scene (rows 1, 2), and static backgrounds are fragmented (row 3). Our method, by contrast, generates a visibly cleaner static point cloud and more coherent, well-separated dynamic objects (e.g., the human in row 4).

This qualitative leap is a direct result of our complete pipeline: (1) Our accurate dynamic masks correctly identify the dynamic regions. (2) Our gradient-aware refinement ensures sharp boundaries, preventing background "floaters". (3) Our 'early-stage masking' strategy effectively prevents this dynamic information from contaminating the deep geometric reasoning layers, allowing for a clean separation.

\subsection{Ablation Studies}
We conduct two crucial ablation studies to validate our core design choices.

\noindent \textbf{Dynamic Mask Estimation Ablation.} In \cref{tab:ablation_mask}, we analyze the components of our mask generation pipeline on the DAVIS dataset.
\begin{itemize}
    \item \emph{Easi3R's logic on VGGT fails.} We first test Easi3R's epipolar-based method on VGGT backbone ($\text{Easi3R}_\text{vggt}$). The performance is poor. This empirically confirms our hypothesis: Easi3R's design is fundamentally incompatible with VGGT's global attention, which does not rely on simple pairwise epipolar geometry.
    \item \emph{Our Gram Similarity mining works.} Our initial mask, generated purely from mining Gram similarities (\textit{w/o refine}), already achieves a very
    strong 59.74 JM. This proves our core motion-mining technique is highly effective and far superior to the Easi3R baseline for this architecture.
    \item \emph{Refinement adds a critical boost.} Adding our gradient-based refinement (\textit{Ours}) provides the final, significant boost to 62.12 JM. This validates its role in removing static regions near motion boundaries and sharpening the final mask.
\end{itemize}

\noindent \textbf{Pose Estimation Ablation.} In \cref{tab:ablation_decoder}, we analyze our early-stage masking strategy for pose estimation on DyCheck. We compare three variants: the baseline \textit{VGGT} (no mask), a naive \textit{Full Mask} (masking dynamic tokens in all layers), and \textit{Ours} (selective 'early-stage' masking).
\begin{itemize}
    \item \emph{Full Masking is harmful.} Naively applying the mask in all layers is worse than using no mask at all (ATE 0.0302 vs. 0.0131). As hypothesized, this pushes the model into an out-of-distribution state and discards too much information.
    \item \emph{Selective Masking is optimal.} Our strategy achieves the best performance (ATE 0.0106). This demonstrates that VGGT benefits from controlled intervention. By only suppressing dynamic tokens in the shallow and mid-level layers, we prevent motion from corrupting the initial geometric features while allowing the deeper layers to operate within their trained distribution.
\end{itemize}

\section{Conclusion}

We present a training-free framework that extends VGGT to 4D scene reconstruction.
We investigate VGGT’s inherent ability to perceive dynamic objects.
By leveraging Gram Similarity signals extracted from VGGT’s attention, we mine and amplify dynamic cues, enabling explicit dynamic disentanglement without relying on any external segmentation modules.
To sharpen dynamic masks, we apply a projection gradient based refinement strategy.
We integrate the refined dynamic masks into VGGT’s early inference, effectively suppressing dynamic interference and improving both pose estimation and geometric reconstruction.
Our method achieves superior quantitative performance and higher-quality visual results.
We hope our findings offer insights for future research on 4D foundation models.

{
    \small
    \bibliographystyle{author-kit/ieeenat_fullname}
    \bibliography{main}
}

\clearpage
\setcounter{page}{1}
\maketitlesupplementary

\section{Empirical Analysis of Attention Layers}
\label{sec:empirical_analysis}

We analyze how the decoder layers of VGGT encode dynamic content. Our findings motivate the specific design choices in our dynamic cue extraction module.

\subsection{Gram Similarity vs. Standard Attention}

\begin{figure}[h]
  \begin{center}
    \includegraphics[width=1.0\linewidth]{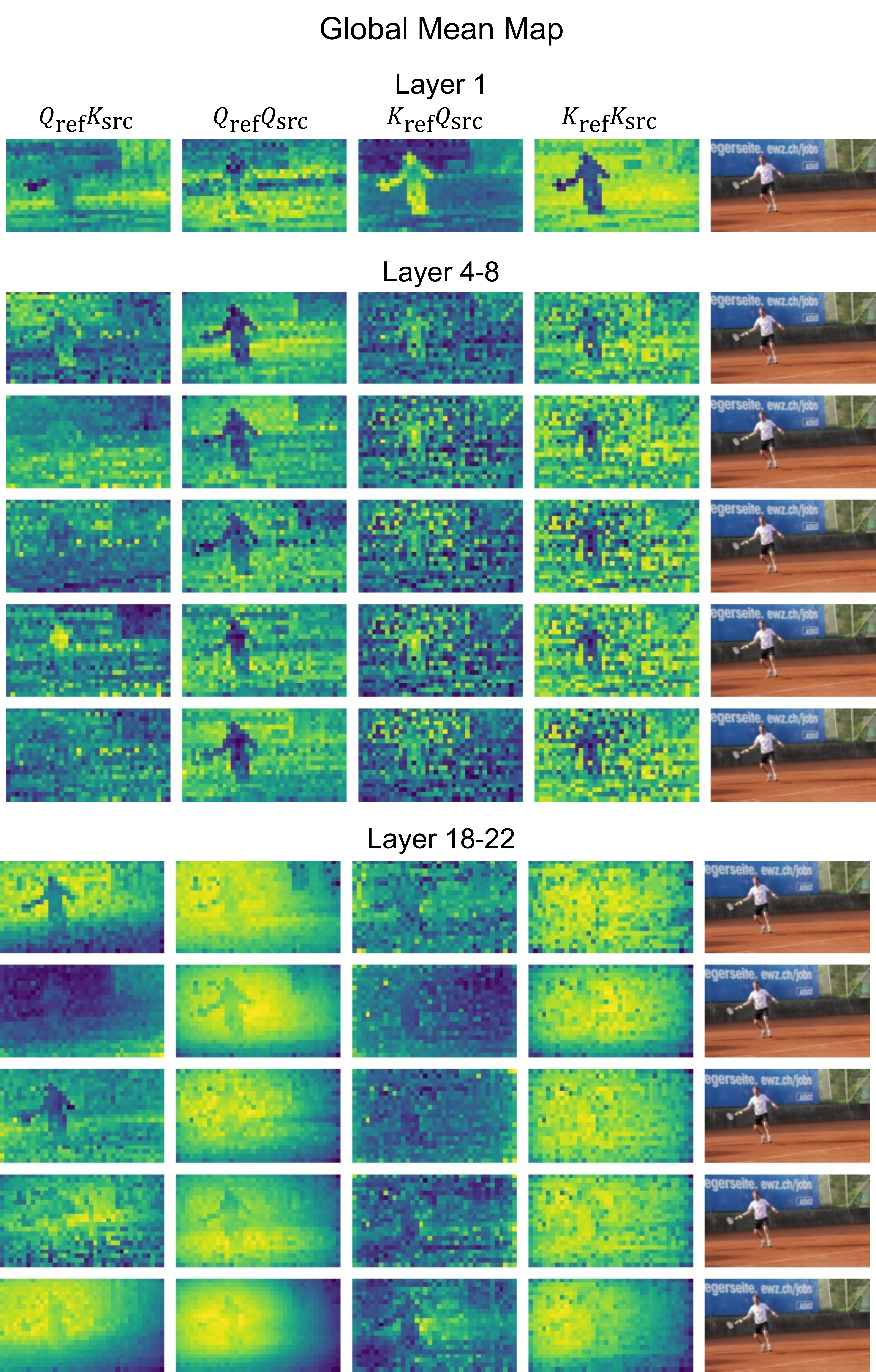}
  \end{center}
  \caption{\textbf{Mean Gram similarity across decoder layers.}
  We visualize Gram similarity maps averaged over a temporal window. The $Q_\text{ref}K_\text{src}$ column shows the standard attention map ($QK^\top$). Note that motion cues are faint in standard attention but distinct in Gram similarities ($QQ^\top, KK^\top$).}
  \label{fig:sup_gram_mean}
\end{figure}

\begin{figure}[h]
  \begin{center}
    \includegraphics[width=1.0\linewidth]{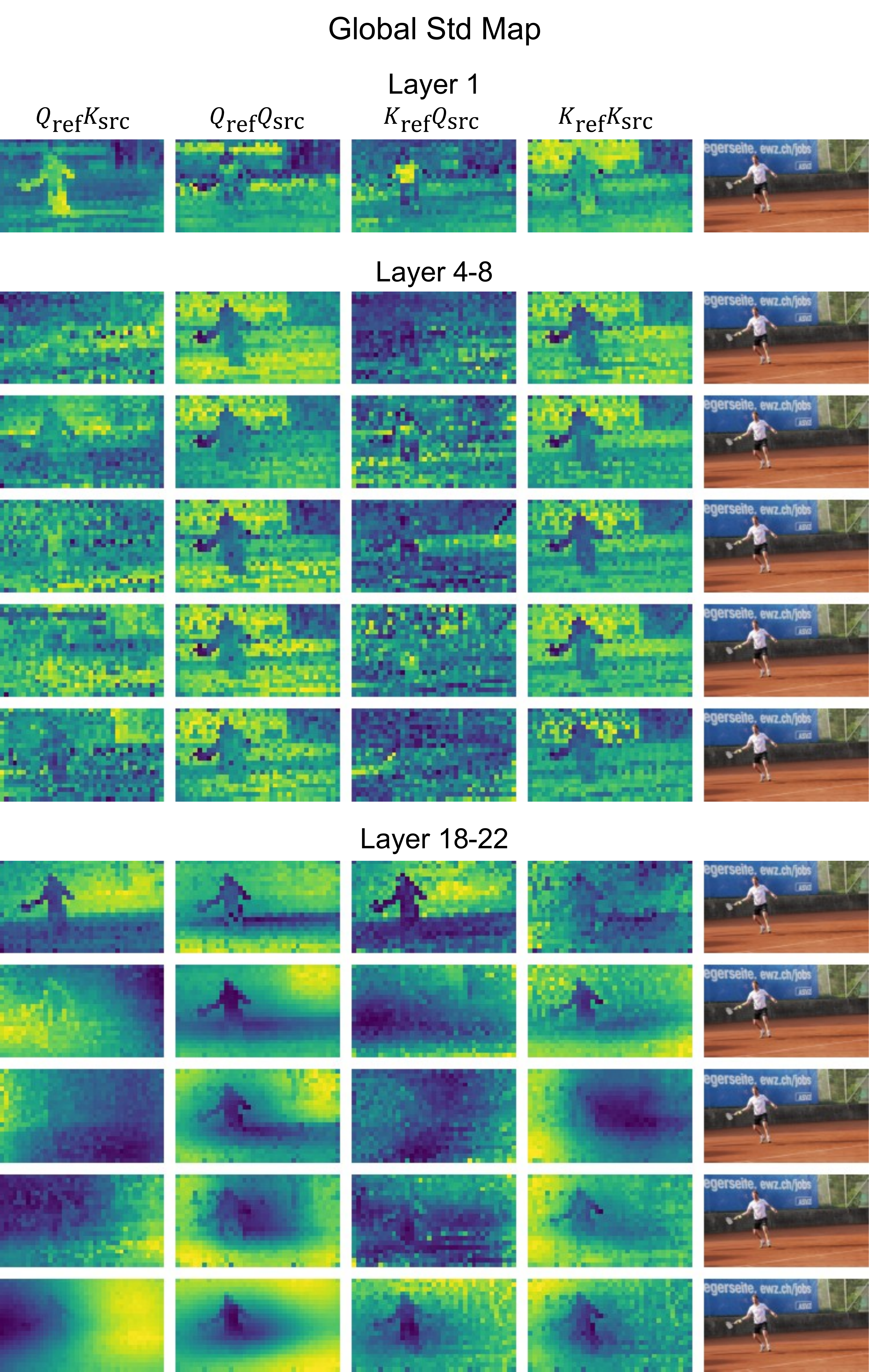}
  \end{center}
  \caption{\textbf{Variance of Gram similarity across decoder layers.}
  We visualize the variance of Gram similarity. Similar to the mean statistics, Gram similarities capture motion variance more effectively than standard attention ($Q_\text{ref}K_\text{src}$).}
  \label{fig:sup_gram_var}
\end{figure}

We first compare standard attention maps with our proposed Gram similarity statistics.
As shown in~\cref{fig:sup_gram_mean,fig:sup_gram_var}, we visualize the statistics (mean and variance) for each decoder layer.
Here, \textit{ref} denotes the reference frame and \textit{src} denotes a source frame sampled from the temporal window.

\noindent\textbf{Standard Attention ($QK^\top$).}
Inspecting the $Q_\text{ref}K_\text{src}$ column reveals that standard attention is dominated by semantic activations. Motion cues are barely visible. This also explains why applying Easi3R to VGGT fails: the strong semantic bias in $QK^\top$ washes out dynamic motion signals.

\noindent\textbf{Gram Similarity ($QQ^\top$ and $KK^\top$).}
In contrast, the Gram similarities (columns $Q_\text{ref}Q_\text{src}$ and $K_\text{ref}K_\text{src}$) make physically dynamic regions salient. This confirms that while motion cues are suppressed in standard attention, they are preserved in the self-similarity of queries and keys.

\subsection{Layer-wise Dynamic Cues}
Based on the Gram similarity patterns, we identify three distinct regimes across the decoder layers:

\begin{itemize}
  \item \textbf{Shallow layers (e.g., layer~1).}
  In the $K_\text{ref}K_\text{src}$ column, the model shows a strong semantic bias. Foreground objects stand out clearly from the background, regardless of their motion state.

  \item \textbf{Middle layers (e.g., layers~4--8).}
  As seen in the $Q_\text{ref}Q_\text{src}$ column, VGGT begins to encode motion variability. The Gram similarities computed from $Q$ vectors over the temporal window sharpen the contrast between truly dynamic regions and the static background.

  \item \textbf{Deep layers (e.g., layers~18--22).}
  In the $Q_\text{ref}Q_\text{src}$ column, spatial priors dominate. This suppresses noisy responses from earlier layers, resulting in sharper and more stable boundaries.
\end{itemize}

\subsection{Ineffectiveness of Camera Token Attention}

\begin{figure}
   \begin{center}
     \includegraphics[width=1.0\linewidth]{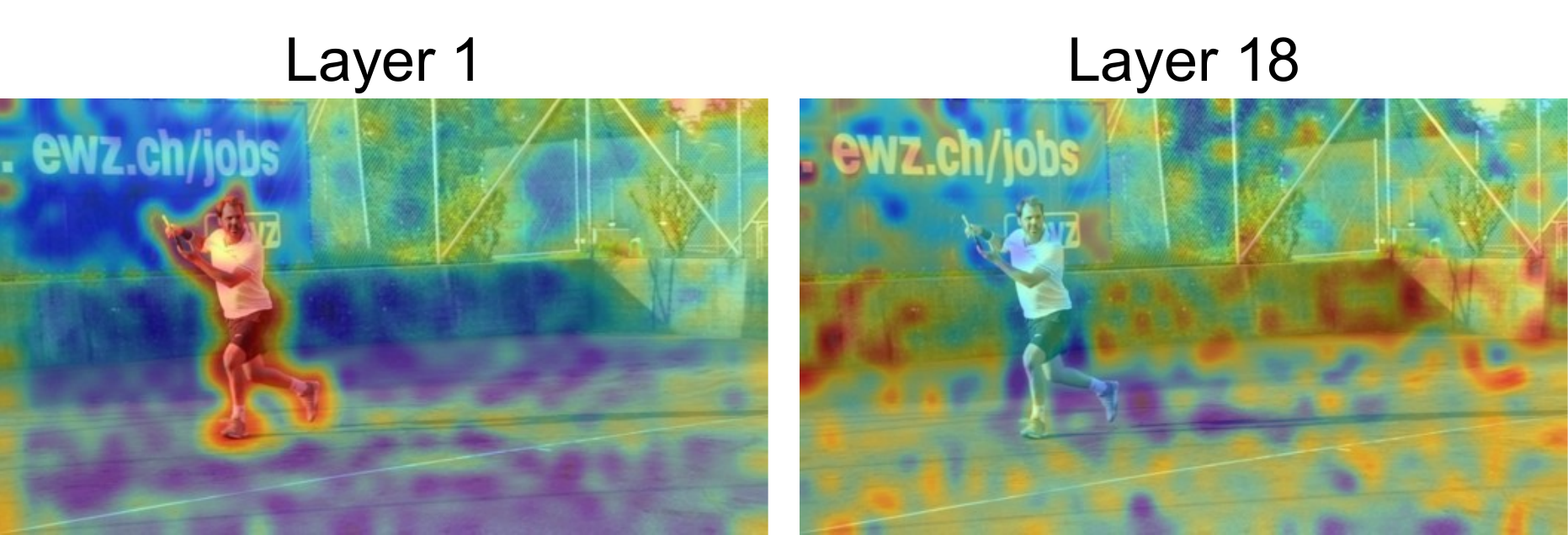}
   \end{center}
   \caption{\textbf{Camera-to-image attention limitations.}
   We visualize attention from the camera token to image tokens. In Layer 18, the camera token fails to suppress the moving actor, treating it similarly to the static background. This confirms that camera tokens are insufficient for robust dynamic masking.}
   \label{fig:sup_cam_attn}
\end{figure}

A potential alternative to our method is relying on the camera token to identify dynamic regions. We test this hypothesis in~\cref{fig:sup_cam_attn}.
While the camera token focuses on dynamic regions in shallow layers (Layer 1), it fails to suppress them in deep layers (Layer 18). The actor's body receives attention comparable to the static background. This indicates that deep camera-token attention is unreliable for dynamic disentanglement, justifying our explicit mining of Gram similarity cues.

\subsection{Explanation of the Design}

\noindent \textbf{Hypothesis on gram similarity.}
We hypothesize that the Gram similarity ($QQ^\top$ or $KK^\top$) serves as a superior amplifier for dynamic cues compared to the standard cross-attention ($QK^\top$). The standard attention mechanism computes interactions between Query and Key vectors, which originate from distinct projection heads to perform semantic alignment. The inherent distributional gap between these heterogeneous vectors tends to overshadow the subtle feature variations induced by object motion, rendering dynamic signals nearly invisible. In contrast, the Gram similarity operates on vectors within the same latent distribution. Without the interference of cross-projection discrepancies, the feature bias introduced by dynamics becomes the dominant source of variance. Consequently, the Gram operation effectively magnifies these intra-distributional differences, making dynamic regions significantly more salient than the static background.

\noindent \textbf{Why camera token is unreliable?}
One might expect the global camera token in deep layers to strictly attend only to static regions for robust pose estimation. However, we observe that Transformer-based foundation models achieve robustness through soft aggregation rather than hard exclusion. The model learns to tolerate a certain degree of dynamic noise (outliers) within its massive context window to solve the global pose optimization. As a result, the camera token does not explicitly `zero out' dynamic regions but rather down-weights them subtly or mixes them into the context. This leads to ambiguous attention maps that are insufficient for generating precise, binary dynamic masks.

\section{Implementation Details}

\subsection{Hyperparameters}
We provide the specific hyperparameters used in our pipeline in \cref{tab:hyperparams}.

\begin{table}[h]
\centering
\footnotesize
\begin{tabular}{l|c}
\toprule
\textbf{Parameter} & \textbf{Value} \\
\midrule
Temporal Window & 6 source frames (stride 2) \\
Layers for $w_{\text{shallow}}$ & Layer 1 \\
Layers for $w_{\text{middle}}$ & Layers 4--8 \\
Layers for $w_{\text{deep}}$ & Layers 19--20 (Var), 18--22 (Mean) \\
Early-stage Masking Layers & Layers 1--5 \\
SOR Neighbors ($k$) & 20 \\
SOR Std. Dev. Mult. ($\sigma$) & 2.5 \\
\bottomrule
\end{tabular}
\caption{\textbf{Hyperparameter settings.} Key parameters for cue extraction and refinement.}
\label{tab:hyperparams}
\end{table}

\subsection{Dynamic Cue Extraction \& Memory Optimization}

As detailed in \cref{sec:empirical_analysis}, we combine cues from three layer groups. We extract $w_{\text{shallow}}$ from Layer 1 for strong semantics. We extract $w_{\text{middle}}$ from Layers 4--8 to capture motion variance. Finally, we extract $w_{\text{deep}}$ from Layers 18--22 to utilize spatial priors for outlier suppression.

To define the threshold $\alpha$, which is used in ~\cref{sec:cue_extraction} to obtain dynamic mask, we use VGGT's ViT backbone to extract image features and apply k-means clustering to group tokens across frames. We verify the dynamic score of each cluster and use Otsu's algorithm to determine the optimal separation threshold.

\noindent\textbf{Memory Optimization.}
VGGT uses PyTorch's Scaled Dot Product Attention (SDPA) for efficiency. Explicitly computing the full Gram matrix $QK^\top$ scales quadratically with token count ($O(N^2)$), which would cause Out-Of-Memory (OOM) errors given the thousands of tokens in multi-view inputs. To resolve this, we compute per-frame Gram similarities in-place within the attention layer. This avoids materializing the massive full matrix and keeps memory usage linear with respect to the number of frames.

\subsection{Mask Refinement}
Raw masks from 3D foundation models often contain outliers (``floaters"). These artifacts degrade the projection-based refinement. We first apply Statistical Outlier Removal (SOR) to the point cloud to filter noise.
Next, we cluster the remaining points and average the projection gradients within each cluster. This aggregation stabilizes the gradient signal, preventing isolated outliers from triggering false positives in the dynamic classification.

\subsection{Early-Stage Masking}
During inference, we apply the dynamic masks only to layers 1--5. We suppress the key ($K$) vectors of dynamic tokens in these layers. This prevents query ($Q$) vectors from attending to dynamic regions early in the network, preserving the geometric consistency of deeper layers.

\section{Additional Experiments}

\subsection{Ablation Study on Dynamic Map Components}
We validate the contribution of each term in \cref{eq:dyn} (Main Paper). As shown in \cref{tab:ablation_dyn}, $w_{\text{shallow}}$ and $w_{\text{middle}}$ provide the primary motion signals. However, performance drops significantly without $w_{\text{deep}}$, confirming its role in suppressing residual outliers.

\begin{table}[h]
\centering
\small
\begin{tabular}{l|cccc}
\toprule
\multirow{2}{*}{\textbf{Method}} & \multicolumn{4}{c}{DAVIS-2016} \\ \cmidrule(lr){2-5}
               & JM↑    & JR↑   & FM↑   & FR↑   \\ \midrule
w/o $w_\text{shallow}$ & 54.15  & 62.44 & 46.43 & 44.27 \\
w/o $w_\text{middle}$  & 56.13  & 57.12 & 44.07 & 41.90 \\
w/o $w_\text{deep}$    & 46.85  & 48.89 & 41.52 & 45.30 \\
w/o refinement  & \underline{59.74} & \underline{73.10} & \underline{50.64} & \underline{58.30} \\ \midrule
Ours           & \textbf{62.12}    & \textbf{76.80} & \textbf{56.04} & \textbf{67.49} \\ \bottomrule
\end{tabular}
\caption{\textbf{Ablation on dynamic mask estimation.} 
We evaluate the contribution of each component. Note that the ``w/o" variants are evaluated before the refinement stage to isolate the impact of the cue extraction.}
\label{tab:ablation_dyn}
\end{table}

\subsection{Zero-shot vs. Trained 2D Segmentation}
We compare our method against FlowSAM, a strong 2D video segmentation baseline. FlowSAM lacks 3D spatial reasoning and typically requires ground-truth supervision (Hungarian matching) for evaluation. For a fair comparison, we adapt FlowSAM to a zero-shot setting.

As shown in \cref{tab:mask_est_v_2d}, our training-free approach outperforms the trained FlowSAM baseline. This demonstrates that leveraging the implicit 3D/4D priors in VGGT yields better temporal consistency than pure 2D video analysis.

\begin{table}[h]
\centering
\small
\begin{tabular}{l|cccc}
\toprule
\multirow{2}{*}{\textbf{Method}} & \multicolumn{4}{c}{DAVIS-2016} \\ \cmidrule(lr){2-5}
                & JM↑                       & JR↑                       & FM↑                       & FR↑                       \\ \midrule
FlowSAM (zero-shot)            & 54.53             & 56.86             & \underline{52.48} & \underline{52.97}                     \\
DAS3R                          & 41.13             & 38.67             & 44.50             & 36.94                     \\
$\text{Easi3R}_\text{dust3r}$  & 50.10             & 55.77             & 43.40             & 37.25                     \\
$\text{Easi3R}_\text{monst3r}$ & \underline{54.93} & \underline{68.00} & 45.29             & 47.30  \\ \midrule
Ours                           & \textbf{62.12}    & \textbf{76.80}    & \textbf{56.04}    & \textbf{67.49}  \\ \bottomrule
\end{tabular}
\caption{\textbf{Dynamic object segmentation comparison.} 
Our method outperforms the trained 2D baseline FlowSAM in a zero-shot setting.}
\label{tab:mask_est_v_2d}
\end{table}

\section{Limitations}

While our method achieves robust 4D reconstruction without training, it has limitations.
First, computing Gram similarities adds computational overhead compared to single-pass inference.
Second, our mask refinement depends on the quality of the initial depth estimates from VGGT. If the backbone misestimates depth (e.g., blending foreground and background), the projection gradients become unreliable.
Finally, our refinement assumes rigid motion for projection checks, which may struggle with highly non-rigid or fluid deformations.
Future work will focus on optimizing efficiency and handling non-rigid dynamics.

\end{document}